# Supervised and Contrastive Self-Supervised In-Domain Representation Learning for Dense Prediction Problems in Remote Sensing

Ali Ghanbarzade, Hossein Soleimani

*Abstract*— In recent years Convolutional neural networks (CNN) have made significant progress in computer vision. These advancements have been applied to other areas, such as remote sensing and have shown satisfactory results. However, the lack of large labeled datasets and the inherent complexity of remote sensing problems have made it difficult to train deep CNNs for dense prediction problems. To solve this issue, ImageNet pre-trained weights have been used as a starting point in various dense predictions tasks. Although this type of transfer learning has led to improvements, the domain difference between natural and remote sensing images has also limited the performance of deep CNNs. On the other hand, self-supervised learning methods for learning visual representations from large unlabeled images have grown substantially over the past two years. Accordingly, in this paper we have explored the effectiveness of in-domain representations in both supervised and self-supervised forms to solve the domain difference between remote sensing and the ImageNet dataset. The obtained weights from remote sensing images are utilized as initial weights for solving semantic segmentation and object detection tasks and state-of-the-art results are obtained. For self-supervised pre-training, we have utilized the SimSiam algorithm as it is simple and does not need huge computational resources. One of the most influential factors in acquiring general visual representations from remote sensing images is the pre-training dataset. To examine the effect of the pre-training dataset, equal-sized remote sensing datasets are used for pre-training. Our results have demonstrated that using datasets with a high spatial resolution for self-supervised representation learning leads to high performance in downstream tasks.

## I. INTRODUCTION

Remote sensing images are becoming one of the most prominent research fields, with various applications including change detection[1-3], land cover classification[4, 5], wildfire detection[6, 7], and climate change[8]. Combining this field with recent advancements in computer vision and deep learning leads to significant performance improvements for each of the mentioned applications.

Dense prediction problems are well-known tasks in computer vision that produce outputs at the pixel level, such as semantic segmentation[9, 10], or box level, such as object detection[11]. Many applications of remote sensing images rely on the high performance of dense prediction tasks. Deep neural networks have demonstrated their capability and effectiveness for dense prediction tasks. However, they require a large amount of labeled training data to learn complex visual features[12, 13]. Labeling images at the box and pixel levels is extremely time-consuming[14]. In the case of satellite images, which are captured from a long distance using various sensors, many different concepts appear, and the object shapes are also different. These differences necessitate domain knowledge, which is rare and inefficient in terms of cost and time[13, 15].

In computer vision, transfer learning is a technique in which a pre-trained convolutional model on a large dataset, such as ImageNet, is used as a starting point for solving other related tasks. This technique reduces the need for labeled data and computation time[15]. In most cases, ImageNet pre-trained models have been utilized as a starting point to solve remote sensing problems[17]. Although ImageNet pre-trained models have often performed well when compared to solving dense prediction problems from scratch, due to the domain gap between ImageNet and remote sensing images, the performance of this type of transfer learning is limited[13]. The first solution which comes into mind is to create a satellite imagery dataset similar to ImageNet, train different convolutional models on it, and then use the obtained in-domain weights to solve other remote sensing tasks. This solution may result in a good performance, but it has some flaws. Firstly, creating a large-scale labeled dataset containing satellite images is extremely expensive and time-consuming. Secondly, pre-training multiple models on such a big dataset requires significant computing resources.

One better solution is hierarchical pre-training which has been proposed in previous studies[17, 18] for image classification problems in remote sensing. Hierarchical Pretraining refers to models that have been pre-trained on datasets that progressively resemble the target task. It includes two pre-training stages: generalist pre-training and specialist pre-training. In generalist pre-training, the convolutional neural network is initially trained on a large data set, such as ImageNet. In the second stage, the weights of the general model are utilized as initial weights for pre-training in other domains, such as remote sensing images. The obtained model can now be used for solving target tasks. The effectiveness of such pre-training is demonstrated by solving some land cover classification problems[18, 19]. In this paper, we aim to apply this method to semantic segmentation and object detection problems, which are significantly more complex than image classification.

Recently, self-supervised learning methods have emerged as the most promising candidates for learning visual representations without the need for large amounts of labeled data. A group of these methods known as contrastive self-supervised learning [20-27] has outperformed other methods for learning general visual representations from natural images. Contrastive learning strategies learn the similarity function between various image perspectives. In these methods, the cost function is defined so as to bring different views of the same image closer in feature space, while separating views originating from different images as much as possible. In computer vision, the pre-trained features from most contrastive self-supervised learning methods have a high degree of generalizability.

In this paper, we use the idea of hierarchical pre-training to solve dense prediction problems in remote sensing images. We select the ResNet50 model pre-trained on ImageNet as the base model, and then train it on the Resisc45[28] and PatternNet[29] datasets, which are created for land cover classification. We have used both supervised and the SimSiam for pre-training in-domain features. We have selected the SimSiam[21] because, despite its simplicity, it does not require negative sampling, momentum encoder, large batch size, or online clustering, so that it requires fewer computational resources. In addition, we have used Resisc45 and PatternNet datasets which have almost equal number of samples to determine the characteristics of the ideal dataset for in-domain pre-training. The generalizability of obtained supervised and self-supervised in-domain features are examined by solving the DeepGlobe Land Cover Classification[30] for semantic segmentation and the Oil Storage Tanks and CGI Airplane problems for object detection.

Our main contributions are as follows:
- We have extracted in-domain visual representations from remote sensing images using both supervised learning and the SimSiam algorithm, one of the most recent contrastive self-supervised learning techniques.
- During the feature extraction phase, ImageNet pre-trained weights are utilized to eliminate the need for larger datasets.
- We have used different ResNet50 models obtained in previous steps as a pre-trained encoder in the DeepLabv3[31] and Faster-RCNN[32] models and we have solved the semantic segmentation and object detection problems respectively.
- We have examined the effect of the pre-training dataset by utilizing equal-sized remote sensing datasets for pre-training.

The remainder of this paper is organized as follows:
In section II, we review the related works. Section III presents the statistics of the selected datasets for each step. Section IV examines the methods used for pre-training in-domain features from remote sensing images. In section V, we solve the downstream tasks, describe the selected models, and demonstrate the results. Finally, we conclude the paper.

## II. RELATED WORKS

### A. Self-Supervised Representation Learning

Self-supervised learning is a sub-branch of unsupervised learning that attracts the attention of researchers all over the world. These methods provide supervisory signals based on the characteristics of the dataset and without the need for human supervision [19]. The self-supervised pre-trained representations are then transferred to the other related supervised downstream tasks to assess their generalizability. The first approach to self-supervised learning includes the design of pretext tasks such as relative position [33], colorization [34], etc. Pretext tasks can be considered proxies for learning the intrinsic patterns and structures of the dataset.

Although researchers have spent a great deal of time and effort designing pretext tasks, these techniques have not yielded significant success in computer vision. For an overview of various pretext tasks designed for different computer vision and natural language processing applications, we refer readers to [19].

In computer vision, pre-trained features from different contrastive self-supervised learning methods [20-27] have recently demonstrated the highest performance in solving various downstream tasks. Most of these methods use data augmentation techniques, such as flipping and cropping, to generate multiple image views [23]. If the inputs stem from the same image, they form positive pairs, and the cost function attempts to bring these views as close in feature space as possible. If the inputs originate from different images, they constitute negative pairs, and the cost function separates them as much as possible in the feature space [35]. Some contrastive learning algorithms require enormous negative samples to capture high-generalizable features [23]. For providing a large number of negative pairs, the PIRL [36] algorithm maintains a memory bank containing the extracted features of every image in the dataset. Therefore, its scalability for real-world applications is limited.

To overcome the former issue, MoCo [24, 25] uses the momentum encoder, and SimCLR [23] uses larger batch sizes. The need for substantial computing resources in the primary contrastive learning algorithms prompted researchers to develop methods that do not need negative samples. For example, SwaV [22] is an example of an online clustering-based algorithm that predicts the code of one view based on the representation of another view of the same image. BYOL [26] is another contrastive learning algorithm that predicts the representation of one view based on the other view and vice versa. Unlike all these methods, SimSiam [21] has eliminated some crucial elements, such as large batch size, momentum encoder, and online clustering. In this paper, we have used the SimSiam algorithm to capture in-domain visual representations from remote-sensing images.

To review different contrastive self-supervised learning methods, we refer the readers to [35, 37, 38]. In the mentioned works, the pros and cons of each method have been analyzed from various perspectives, making them useful for comparing different approaches.

### B. Representation Learning for Remote Sensing Imagery

The domain difference between remote sensing and ImageNet images can be addressed by in-domain pre-training. Reviewing recent literature makes it possible to imagine two distinct approaches for obtaining general visual representations from remote sensing images: supervised and self-supervised learning. Therefore, using supervised or self-supervised learning strategies, one can pre-train the models on satellite images before transferring the pre-trained weights to other similar tasks. As an example of the supervised approach, in [17], after pre-training the ResNet50 model on satellite imagery datasets, the obtained weights are utilized to solve various land cover classification tasks. Similar to previous work, in [15], the ResNet50 model is pre-trained from scratch on satellite images, and the authors transferred the obtained weights to similar tasks. The results of these papers demonstrated that the in-domain pre-trained models have comparable or even better results than the ImageNet pre-trained counterparts on various remote sensing tasks and datasets.

Recently, some researchers have examined the quality of pre-trained features obtained from different self-supervised learning methods on overhead imageries. For instance, in [41], various ways of self-supervised pre-training, such as image inpainting, content prediction, and instance discrimination, have been investigated. In [40], a new pretext task was defined which uses the information of invisible channels from muti-spectral remote sensing images to predict the contents of the RGB channels. For the first time, the authors of Tile2Vec [42] have utilized the concept of contrastive self-supervised learning to capture in-domain features from remote-sensing images. This algorithm employs a form of the triplet loss function to bring the anchor tile closer to the neighbor tile while moving away from the distant tile in the feature space simultaneously.

In [18], the authors introduced hierarchical pre-training which significantly reduced the convergence time, the need for computing resources, and the number of pre-training samples. Hierarchical Pretraining refers to models pretrained on datasets that are progressively more similar to the target data. The authors utilized the ImageNet pre-trained MoCo [24] algorithm as the base model. Then, using the MoCo algorithm, the obtained weights are optionally pre-trained on another dataset similar to the target dataset. Finally, the resulting model has been fine-tuned on the target datasets. Recently, [16] has reviewed various self-supervised learning methods used for remote sensing imagery analysis, including RGB, Hyper-spectral, etc. For additional information, please refer to [16].

In this paper, for the first time, we have investigated the effectiveness of hierarchical pre-training to address dense prediction problems in remote sensing images. Firstly, as the base model, we have used the ResNet50 model, which is pre-trained on the ImageNet dataset. Secondly, we fine-tuned the base model by supervising and self-supervised approaches on different overhead datasets. Finally, we have used the resulting model as a pre-trained backbone in semantic segmentation and object detection models to solve high-level dense prediction tasks.

### C. Semantic Segmentation of DeepGlobe Land Cover Classification

Deep CNNs have shown superior performance in solving various semantic segmentation problems. For semantically segmenting objects, models such as UNet, SegNet, DeepLab series, etc., have been developed [43]. Since DeepGlobe Land Cover Classification is one of the tasks addressed in this article, we provide a brief overview of the prior research related to this dataset. Models such as FCN and other encoder-decoder-based structures have been widely used to accomplish the task [44-52]. For instance, in [45], remote sensing images were semantically segmented using U-Net architecture. Similarly, in [47], researchers employed the DeepLabv3+ model to solve the DeepGlobe Land Cover Classification task.

In [13], the image inpainting pretext task is applied to the flow dataset to learn in-domain visual representations. Different satellite image semantic segmentation problems, such as scene parsing, road extraction, and land cover estimation, have been assigned pre-trained weights.

## III. DATASETS

We have used two types of remote sensing datasets for our experiments. The first category consists of medium-sized land cover classification datasets, which have been utilized to pre-train in-domain visual representations. The second category consists of datasets specific to dense prediction problems in satellite images and is used to evaluate the pre-trained features. All of our experiments are conducted on RGB remote-sensing datasets.

### A. Pre-training Datasets

Choosing an appropriate remote-sensing dataset for pre-training general features is one of the most vital factors in the performance of models on downstream tasks. To examine the impact of this factor, we have selected two datasets with comparable sample sizes and the number of classes. In the subsequent section, we have presented the details of each dataset:

**PatternNet**[29]**:** PatternNet consists of 38 classes with 800 images per class. Therefore, this dataset contains 30,400 samples. The image resolution of this dataset is 256 by 256 pixels. In addition, this dataset has a spatial resolution between 0.06m and 4.96m.

**NWPU-RESISC45**[28]**:** This dataset consists of 31,500 images classified into 45 categories. This dataset contains images with a spatial resolution between 0.2m and 30m per pixel for many samples. The image resolution of this dataset is 256 by 256 pixels. Table I summarizes the characteristics of both datasets:

TABLE I
Statistics of the pre-training datasets.

| Dataset | Image size | # Images | Classes | Resolution (m) |
|---|---|---|---|---|
| **Resisc45** | 256x256 | 31.5k | 45 | 0.2-30m |
| **PatternNet** | 256x256 | 30.4k | 38 | 0.06-4.96m |

### B. Dense Prediction Datasets

Semantic segmentation and object detection are selected as downstream tasks to evaluate pre-trained features. The selected datasets are as follows:

**DeepGlobe Land Cover Classification** [30]: This dataset consists of 803 RGB images with pixel-level labels. The size of each image is 2448x2448 pixels, and the spatial resolution is 0.5m. In addition, there are seven distinct classes, including background. Each category is color-coded with a unique color. We have demonstrated the complete information of the dataset in Table II.

TABLE II
Distributions of the different classes in the DeepGlobe land cover classification dataset(completely imbalanced) [30].

| Classes | Pixel count | Proportion | Color code |
|---|---|---|---|
| Urban_land | 642.4M | 9.35% | [0, 255, 255] |
| Agriculture_land | 3898.0M | 56.76% | [255, 255, 0] |
| Rangeland | 701.1M | 10.21% | [255, 0, 255] |
| Forest_land | 944.4M | 13.75% | [0, 255, 0] |
| Water | 256.9M | 3.74% | [0, 0, 255] |
| Barren-land | 421.8M | 6.14% | [255, 255, 255] |
| **Unknown** | 3.0M | 0.04% | [0, 0, 0] |

In the above table, The Pixel count and Proportion columns show the number and percentage of pixels of each class in the dataset, respectively. Accordingly, the dataset is class imbalanced, and identifying unknown, water, barren land, and urban land are extremely arduous.

**Oil Storage Tanks (OST):** It contains 10,000 images, of which 1,829 have box-level labels in three classes. Each image has 512x512 pixels. It was recently proposed for one of the Kaggle competitions but has not appeared in recent publications. In our experiments, we have used 1500 images for fine-tuning the pre-trained features and the remaining 329 images for evaluation.

In Table III, we have listed the number of labeled samples for each of the three classes for either training or evaluation sets:

TABLE III
The number(percentage) of annotated samples for each class in the train and test sets. tank cluster is a rare concept.

| **Class** | **Train_Samples (%)** | **Test_samples (%)** |
|---|---|---|
| Tank | 2446(32.9%) | 596(36.5%) |
| Tank Cluster | 158(2.2%) | 31(1.9%) |
| Floating Head Tank | 4815(64.9%) | 1005(61.6%) |
| Total samples | 7419 | 1632 |

According to the above table, tank cluster samples are scarce, and identifying the tank clusters in images is tedious work.

**CGI planes in satellite imagery with bounding box:** It consists of 500 synthetic overhead images and contains 1000x700 pixels per image. In our experiments, we have used 400 samples to train the object detection model and the remaining 100 images to evaluate models.

## IV. METHOD

Our objective is to extract meaningful features from satellite images using both supervised and self-supervised techniques, and then to use the learned weights as initial weights for semantic segmentation and object detection. Consequently, the employed method has two phases, which are:
1. Pre-training ResNet50 model both supervised and self-supervised on Resisc45 and PatternNet datasets
2. Transferring pre-trained weights in the previous step to the backbone of semantic segmentation and object detection models, and solving dense prediction problems.

In the pre-training phase, the ImageNet weights are adjusted; thus, the pre-trained model is also utilized in the pre-training. All codes were written utilizing the PyTorch and PyTorch Lightning[53] frameworks and executed on an Ubuntu system with a QuadroP6000 GPU.

A. Pre-Training of In-Domain Visual Representation

To obtain visual representations from remote sensing images, we have considered the ResNet50 model as a backbone and pre-trained this model both in a supervised and by using the SimSiam algorithm, which is based on contrastive self-supervised learning, on Resisc45 and PatternNet datasets. During pre-training, we use ImageNet pre-trained ResNet50. The pre-training datasets have almost the same number of samples, but they are different in terms of spatial resolution. We have selected these two datasets because our second goal is to examine the effect of the pre-training dataset on the final performance.

**Supervised pre-training of in-domain visual representations:** To obtain supervised in-domain visual representations, we trained the ResNet50 model with 90% of the samples from both the Resisc45 and PatternNet datasets and evaluated the learned features with the remaining 10%. The objective of this article is not to solve classification problems. Therefore, we have used only 10% of the data as a test set to determine the right direction during pre-training. In our experiments, with a batch size of 120, we have pre-trained the model for 100 epochs. Additionally, we have utilized OneCycleLR as a learning rate scheduler and Adam as an optimizer. In table IV, we have reported the global accuracy:

TABLE IV
Accuracy (%) of ResNet50 model on validation set (10% of each dataset)

| **Dataset** | **Global Accuracy (%)** |
|---|---|
| PatternNet | 99.97 |
| Resisc45 | 97.20 |

By conducting this experiment, we have produced two pre-trained models called Sup-Resisc45 and Sup-PatternNet, which were pre-trained on Resisc45 and PatternNet, respectively.

**Self-supervised pre-training of in-domain visual representations using the SimSiam algorithm:** We have selected the SimSiam algorithm for pre-training the general features in a self-supervised way because its pre-trained features for natural images are highly generalizable. Furthermore, since it does not require a negative sample, a larger batch size, and a memory bank, it is optimal in terms of computation and resource requirements. We have used all images of Resisc45 and PatternNet for pre-training in-domain features by SimSiam.

The encoder of the SimSiam algorithm consists of a backbone, a projection head, and a prediction head [21]. In our experiments, ResNet50 serves as the network's backbone. We have trained this model on two datasets, PatternNet and Resisc45, for 400 epochs. During pre-training, we utilized the SGD optimizer with a batch size of 128 and a base learning rate of 0.05, along with the MultiStepLR scheduler. We have set the weight decay to 0.00001 and the SGD momentum to 0.9. By conducting this experiment, we have produced two additional pre-trained models called Sim-Resisc45 and Sim-PatternNet, which were pre-trained on Resisc45 and PatternNet, respectively.

## V. DENSE PREDICTION TASKS

A. Semantic Segmentation

**Semantic segmentation using the DeepLabV3 algorithm:** We have re-implemented the DeepLabV3, a well-known semantic segmentation model, to make supervised and self-supervised transfer learning feasible. To reduce the semantic segmentation model's convergence time, we have set the overall output stride to 32. Additionally, the last fully-connected layers of the decoder part of the DeepLabv3 are modified, reducing the number of parameters by approximately 0.6m compared to the default model.

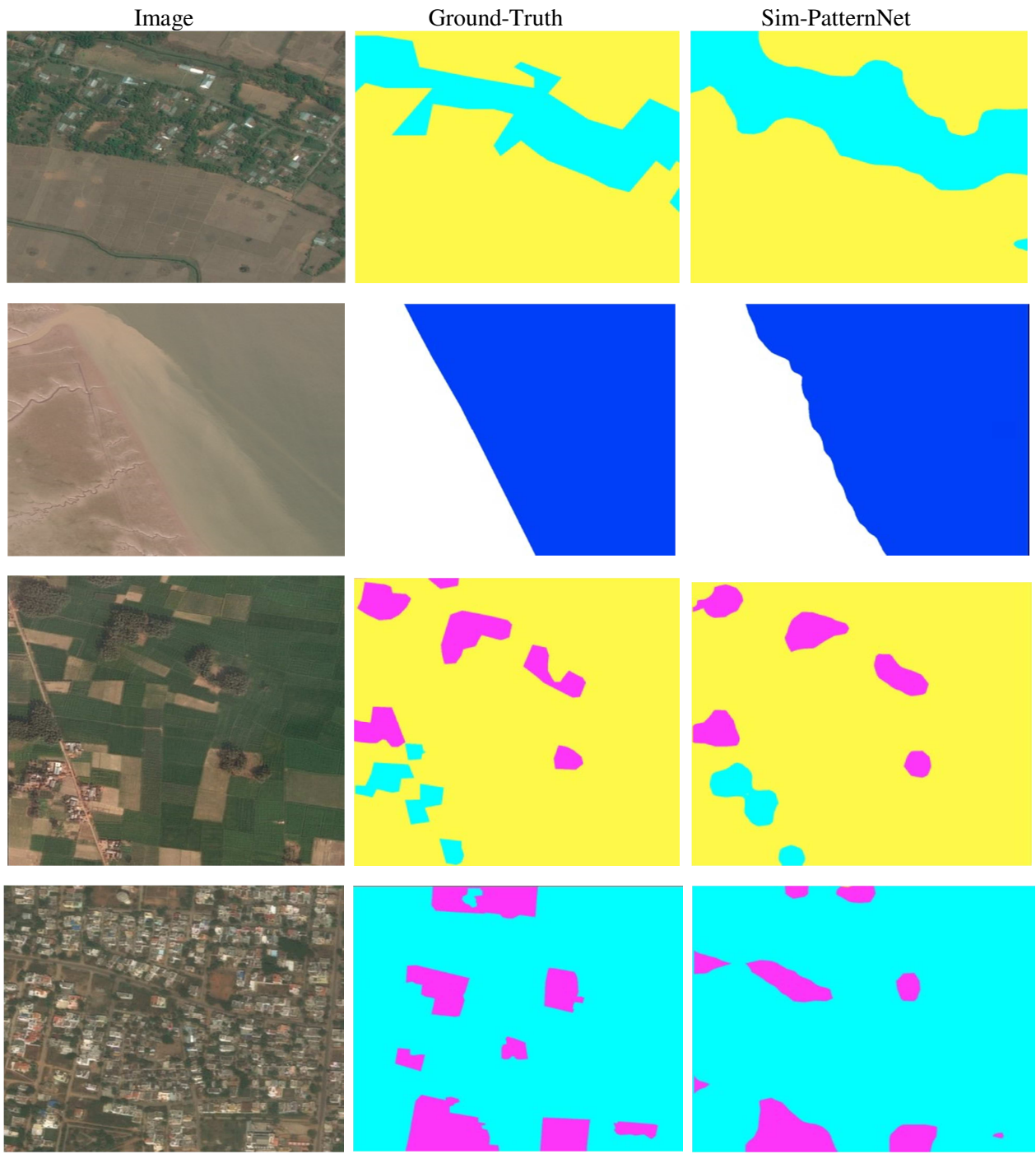

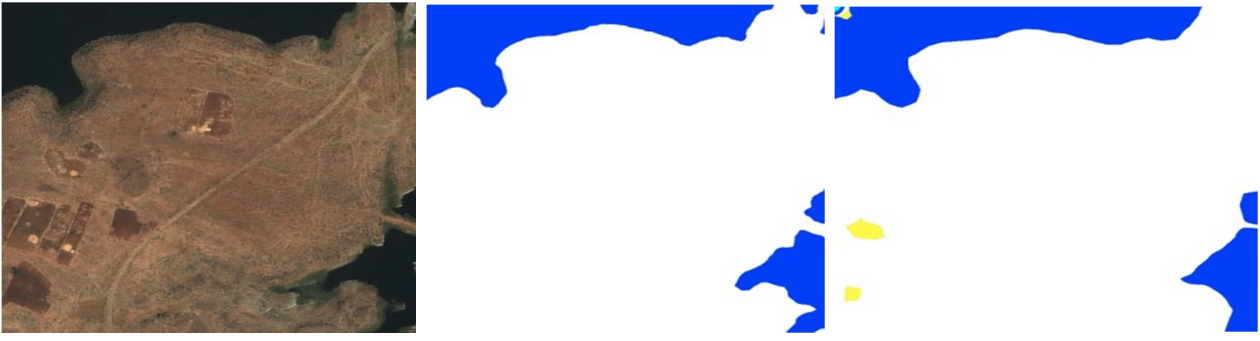

**Figure 1.** Predictions of the model on some of the challenging test samples. First column contains the original images, second column contains the ground truth masks, and third column contains Sim-PatternNet predictions.

We have transferred supervised and self-supervised pre-trained weights obtained in the previous section to the semantic segmentation task. In one of our experiments, as a base model, we replaced the DeepLabV3's backbone with the ImageNet pre-trained ResNet50 and named it Base-Model. Under these circumstances, we have obtained five different DeepLabV3 models, including Base-Model, Sup-Resisc45, Sup-PatternNet, Sim-Resisc45, and Sim-PatternNet. Therefore, ImageNet pre-trained, supervised Resisc45 pre-trained, supervised PatternNet pre-trained, self-supervised Resisc45 pre-trained, and self-supervised PatternNet pre-trained models serve as our backbones.

**Evaluation Metrics:**

**mIOU:** The mean Intersection Over the Union, also known as the Jaccard Index, is used to evaluate semantic segmentation models. This metric is very explicit and fully determines segmentation model performance. If the value of mIOU for a model is equal to one, it indicates that the model predicts accurately [54].

**Pixel Accuracy (PA):** Pixel accuracy is another metric used to evaluate semantic segmentation models. It is the proportion of correctly classified image pixels. This metric is defined and calculated independently for each class, and the resulting values are then averaged [54].

**F1-score (f1):** Another metric to evaluate the semantic segmentation model is the f1-score. This metric is calculated based on precision and recall, which are defined as follows:

$$\text{precision} = \frac{TP}{TP+FP} \quad , \quad \text{Recall} = \frac{TP}{TP+FN} \quad (1)$$

In the above equations, TP, FP, and FN represent True Positive, False Positive, and False Negative, respectively. The combination of Precision and Recall is called the f1-score and is defined as follows:

$$f1-\text{score} = 2 \times \frac{\text{Precision} \times \text{Recall}}{\text{Precision}+\text{Recall}} \quad (2)$$

The closer the f1-score of an algorithm is to one, the better the performance of the segmentation model [19, 54].

**Semantic segmentation with supervised and self-supervised pre-trained models:** To evaluate the performance of the pre-trained models, we have fine-tuned them using the DeepGlobe Land Cover Classification dataset. In all our experiments, we used 80% of the data (642 images) for training and 20% (161 images) to evaluate the segmentation model. During training, we utilized the Adam optimization algorithm with a batch size of 4 and a weight decay of 0.0001. As data augmentation, we have applied random horizontal and vertical flipping and cropping from 2448×2448 to 1024×1024. We trained each of the models on the train set of the DeepGlobe Land Cover Classification dataset for only five epochs. The obtained results for each of the pre-training methods can be seen in Table V:

TABLE V
Comparison of the different pre-training methods. The Sim-PatternNet model has improved the MIOU by 2.1% compared to the base-model.

| Method | PA (%) | f1(%) | mIOU (%) |
|---|---|---|---|
| **Base-Model** | 95.32 | 85.12 | 77.56 |
| **Sup_Resisc45** | 95.68 | 85.73 | 77.75 |
| **Sup_PatternNet** | 95.90 | 86.34 | 78.50 |
| **Sim_Resisc45** | 95.78 | 85.45 | 76.37 |
| **Sim_PatternNet** | **96.20** | **87.38** | **79.66** |

According to the above table, we can draw several significant conclusions regarding in-domain pre-training on remote sensing images and its fine-tuning on semantic segmentation tasks. 1) For pre-trained models using the SimSiam algorithm, even though the class diversity and the number of Resisc45 pre-training samples are more than PatternNet, the pre-trained model on the PatternNet performs better. Pixel spatial resolution is the primary distinction between these two datasets. Table 1 indicates that Resisc45 and PatternNet are both multi-resolution datasets. Assuming that the spatial resolution distribution of all pixels is uniform, the average pixel resolution of Resisc45 and PatternNet is 15.1m and 2.51m, respectively. Therefore, on average, PatternNet spatial resolution is significantly higher than Resisc45. This factor makes the pre-trained features from PatternNet more precise than those from Resisc45. 2) In the case of supervised in-domain pre-trained models, the PatternNet model outperforms the Resisc45 model, confirming the former conclusion. 3) Inspecting the performance of the resisc45 pre-trained models reveals that the supervised model outperforms the self-supervised counterpart. In contrast, for PatternNet pre-trained models, the SimSiam pre-trained model performs better. Generally, in self-supervised pre-training from remote sensing images, it is preferable to select datasets with higher spatial resolutions. In Table VI, we compare the results obtained from

our best Sim-PatternNet model to those of other relevant works:

TABLE VI
Comparison of our best results with other methods. Self-supervised pre-training on PatternNet leads to state-of-the-art results.

| References | mIOU (%) | PA (%) |
|---|---|---|
| [45] | 42.80 | ---- |
| [47] | 51.00 | ---- |
| [50] | 65.2 | ---- |
| [52] | ---- | 80.49 |
| [51] | 66.1 | 84 |
| [44] | 67.87 | 86.58 |
| [49] | 75.6 | ---- |
| **Ours** | **79.66** | **96.20** |

According to the above table, hierarchical pre-training on ImageNet and then on relevant satellite datasets produces features with high generalization power. Compared to previous works, transferring the Sim PatternNet pre-trained weights to the DeepLabV3 for solving the DeepGlobe Land Cover Classification yielded a state-of-the-art result.

Figure 1 demonstrates predictions made by the Sim-PatternNet model on challenging test samples. The first column in this figure contains the test image, the second column contains the corresponding ground truth, and the third column displays the prediction of the DeepLabV3 model with the Sim-PatternNet pre-trained backbone. According to Figure 1, the presented model can accurately classify pixels comprising a small percentage of the dataset, such as water, barren land, and urban land.

B. Object Detection

**Object Detection with Faster-RCNN model:** We have used the Faster-RCNN with the FPN implemented in Detectron2 to solve object detection problems. Again, we have distinguished different models with Base-Model, Sup-Resisc45, Sup-PatternNet, Sim-Resisc45, and Sim-PatternNet. Various evaluation metrics, including AP, AP50, AP75, APs, APm, and APl [55] are used for comparing the performance of different models.

**Oil Storage Tanks:** First, we have fine-tuned each model with 5000 iterations and a batch size of 2 on the training part of the dataset. The initial learning rate and warm-up iterations are 0.01 and 500, respectively. Then, the obtained object detection models are evaluated on the test samples. Finally, we have reported different metrics to illustrate the advantages and disadvantages of each pre-training method. Table VII, demonstrates the obtained results for each of the models:

TABLE VII
Comparison of different pre-training methods in the oil storage tanks detection (a Kaggle competition). The Sim-PatternNet model has improved the AP by 2.65% compared to the base-model.

| Method | AP | AP50 | AP75 | APs | APm | APl |
|---|---|---|---|---|---|---|
| **Base-Model** | 37.60 | 46.37 | 42.77 | 30.87 | 41.15 | 75.58 |
| **Sup-Resisc45** | 38.02 | 46.95 | 42.74 | 29.24 | 41.11 | **78.04** |
| **Sup_PatternNet** | 39.46 | 49.92 | 45.49 | 25.46 | **44.67** | 77.41 |
| **Sim-Resisc45** | 37.84 | 48.35 | 43.23 | 29.9 | 42.10 | 74.38 |
| **Sim-PatternNet** | **40.25** | **50.31** | **46.83** | **34.04** | 40.89 | 75.16 |

According to the results in the above table, in-domain pre-training of features using supervised and self-supervised approaches surpass the ImageNet pre-trained model. Considering AP as a representative candidate for evaluating the performance of object detection models, the same conclusions for the semantic segmentation task can be valid here. Again, Sim-PatternNet outperforms all other models. Table IV reveals that a large portion of the labeled samples in the dataset belongs to the Tank and Floating Head Tank. In contrast, the Tank Cluster is less frequent. By taking this concept into account, we have shown the AP of various models for each class in Table VIII:

TABLE VIII
Comparison of different pre-trained models for each class (AP of each class). Self-supervised pre-trained models detect rare concepts better.

| Method | Tank | Tank Cluster | Floating Head Tank |
|---|---|---|---|
| Base-Model | 52.46 | 0.00 | 59.26 |
| Sup-Resisc45 | 52.10 | 0.17 | 61.80 |
| Sup-PatternNet | 53.95 | 0.26 | **64.19** |
| Sim-Resisc45 | 51.36 | 1.19 | 60.98 |
| Sim-PatternNet | **54.26** | **3.56** | 62.94 |

| Annotated Image | Base-Model | Sim-PatternNet |
|---|---|---|

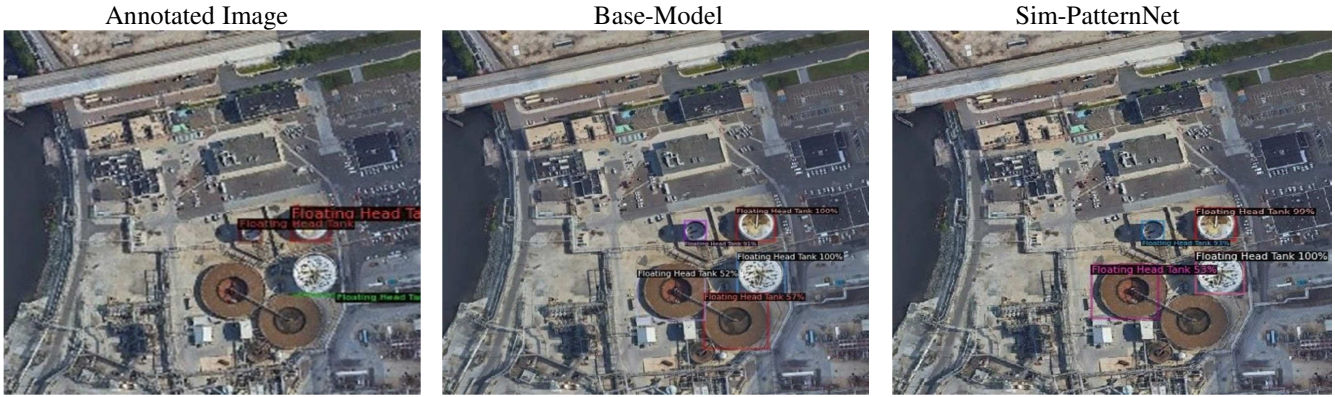

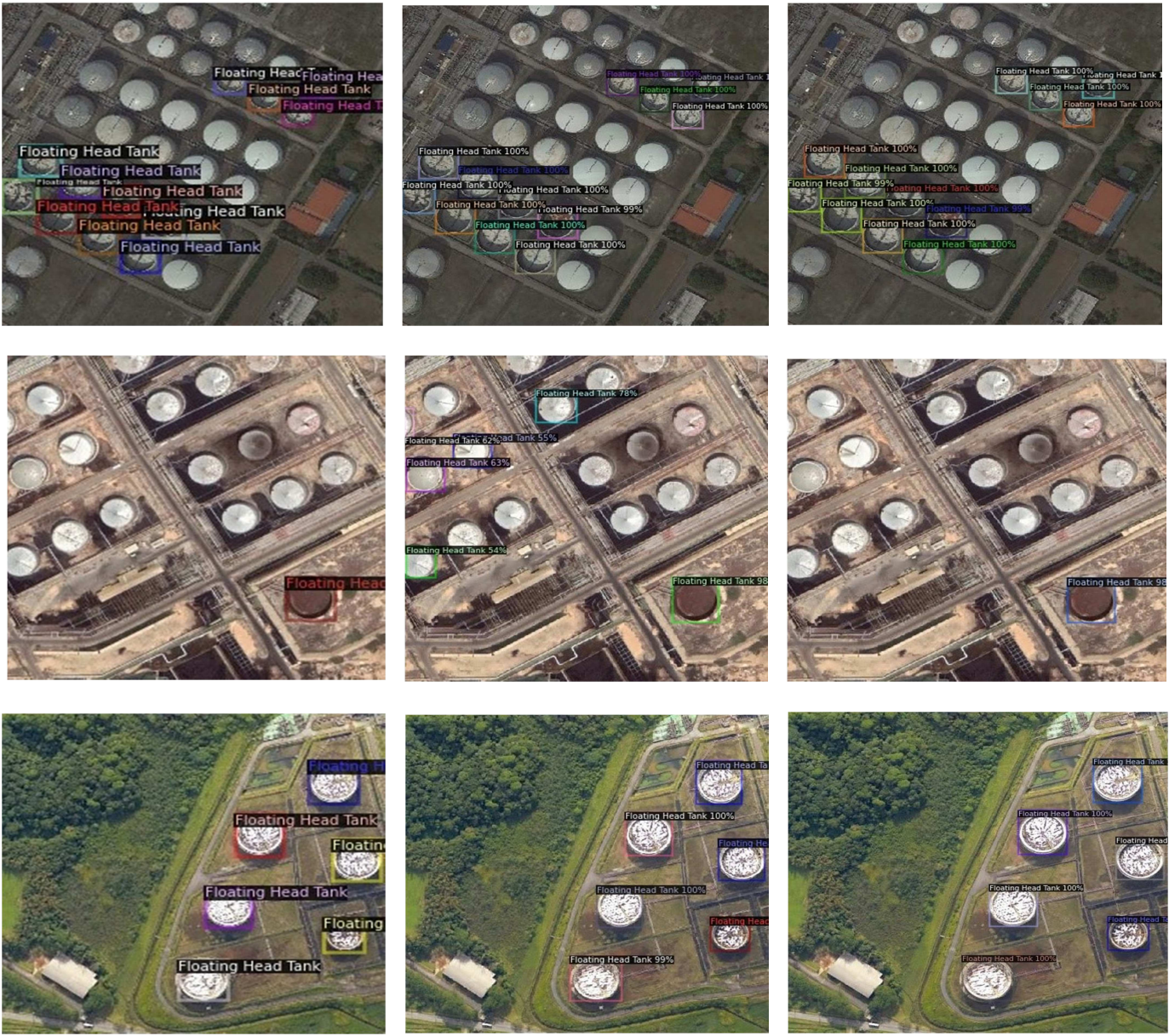

**Figure 2.** Predictions of the models on some of the challenging test samples. First column contains the original images, second column contains the Base-Model results, and third column contains the Sim-PatternNet predictions.

Tank Cluster is only 2.1% of all labeled objects in the dataset. Therefore, it is difficult to identify the bounding boxes of this class. As shown in Table VIII (second column), self-supervised pre-trained models on either PatternNet or Resisc45 perform better than other models when the number of labeled samples is limited. In Figure 2, the first column shows challenging images of the test set and their corresponding label. The second and third columns illustrate the prediction of the Faster-RCNN model with the Base-Model and Sim-PatternNet as the backbone, respectively. We can conclude that by utilizing the pre-trained Sim-PatternNet as the backbone of the Faster-RCNN, the object detection model predicts more precise bounding boxes for each object in all images.

**CGI Planes Detection:** This dataset is selected because all its images are synthetic (computer generated), so comparing different models' performance on this dataset is worthwhile. We used 80% (400 images) of the dataset for training and 20% (100 images) for model evaluation. We have fine-tuned each model on the train part for 2,000 iterations with a batch size of two. The initial learning rate and the warm-up iterations are 0.015 and 300, respectively. The obtained results for different models can be seen in Table IX.

TABLE IX
Comparison of different pre-training methods in the CGI-airplane detection (a Kaggle competition). the Sim-PatternNet model has improved the AP by 1.47 compared to the base-model.

| Method | AP | AP50 | AP75 | APs | APm | APl |
|---|---|---|---|---|---|---|
| **Base-Model** | 78.66 | 97.00 | 92.51 | 50.19 | 78.34 | 86.80 |
| **Sup-Resisc45** | 78.48 | 97.02 | 92.65 | **68.33** | 78.41 | 82.88 |
| **Sup-PatternNet** | 78.96 | 97.01 | **93.61** | 65.92 | 78.39 | 86.35 |
| **Sim-Resisc45** | 77.99 | 97.00 | 93.60 | 66.78 | 77.78 | 82.64 |

| Sim-PatternNet | 80.13 | 97.94 | 92.57 | 58.66 | 79.99 | 87.00 |

According to the second column of the preceding table, the self-supervised pre-trained model on the PatternNet dataset outperforms other models when fine-tuning the pre-trained weights on the airplane detection dataset. Therefore, the previous conclusions are consistent with this experiment.

In one of our experiments, we used the Faster-RCNN pre-trained on the COCO2017 dataset (COCO-Model) as the initial model, then we fine-tuned COCO-Model weights on CGI airplane detection. In Table X, we have compared the results of the Base Model, Sim-PatternNet, and COCO-Model.

TABLE X
The performance comparison of our best pre-trained model with the result of fine-tuning the COCO 2017 pre-trained model.

| Method | AP | AP50 | AP75 | APs | APm | APl |
|---|---|---|---|---|---|---|
| Base-Model | 78.66 | 97.00 | 92.51 | 50.19 | 78.34 | 86.80 |
| Sim_PatternNet | 80.13 | 97.94 | 92.57 | 58.66 | 79.99 | 87.00 |
| COCO-Model | 80.54 | 98.01 | 94.60 | 65.36 | 80.27 | 88.20 |

According to the above table, the pre-trained model on COCO2017 performs better than other models in all reported evaluation metrics. However, the Sim-PatternNet model has its merits because we must consider that the SimSiam algorithm only pre-trains the backbone of the Faster-RCNN. In contrast, the COCO-Model is an entire Faster-RCNN pre-trained model. In addition, during pre-training, self-supervised pre-training does not need human supervision, while COCO-Model uses precise human labels at the box level. In Figure 3, we have demonstrated the performance of pre-trained ImageNet and Sim-PatternNet models on some challenging test samples. Figure 3 reveals that the Sim-PatternNet model outperforms the Base Model for synthetic aerial images.

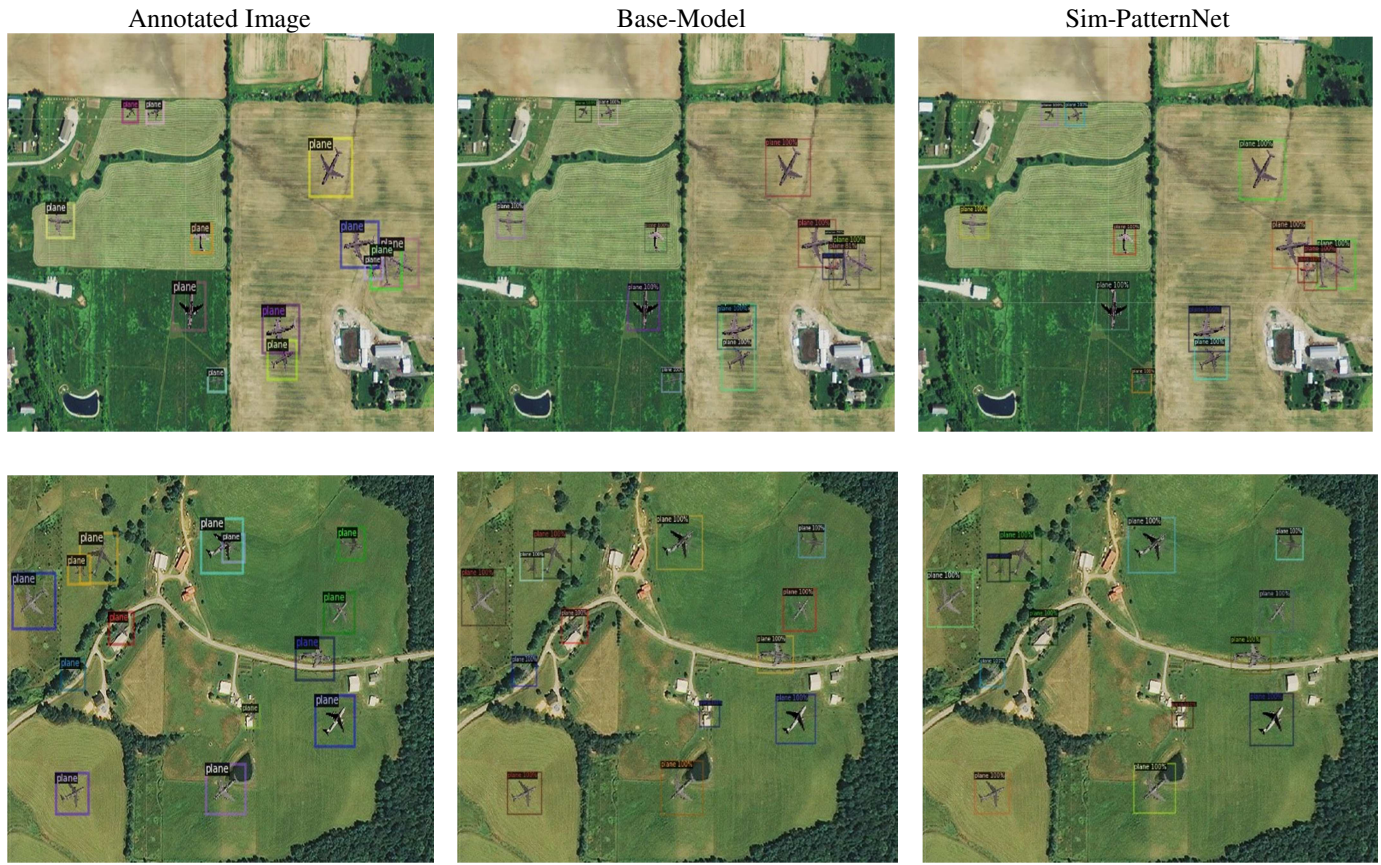

Annotated Image | Base-Model | Sim-PatternNet

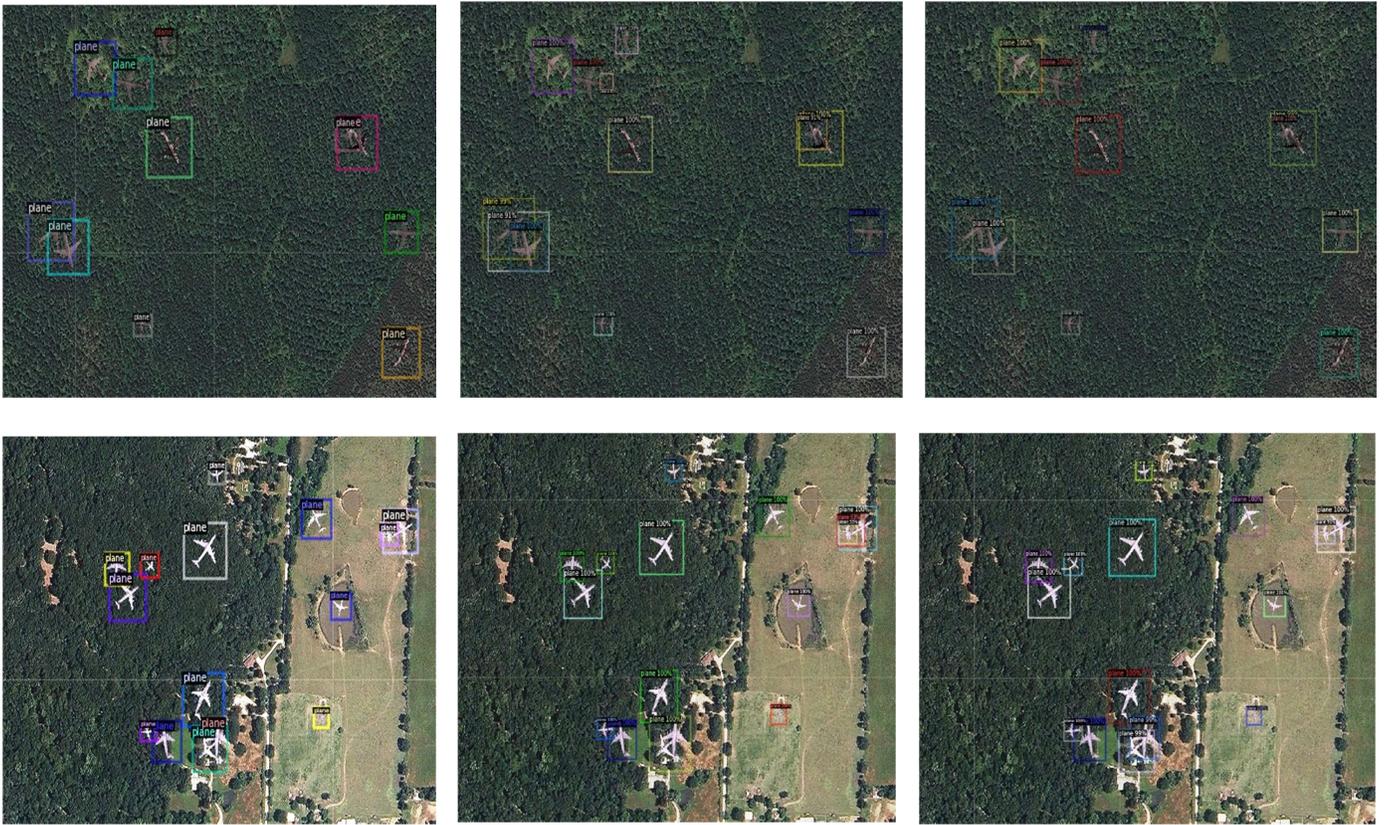

Figure3: Predictions of the models on some of the challenging test samples. First column contain the original images, second column contain the Base-Model results, and third column contain Sim-PatternNet predictions.

## VI. CONCLUSIONS

ImageNet pre-trained models have long been dominant for transfer learning in solving various remote sensing tasks. However, the domain difference between ImageNet and remote sensing images limits the performance of this type of transfer learning, especially for dense prediction problems. One possible solution is to pre-train in-domain features from remote sensing datasets. In this paper, we investigated the generalizability of pre-trained in-domain features using supervised and contrastive self-supervised learning approaches. During pre-training, we pre-trained the ResNet50 model on Resisc45 and PatternNet datasets. Although many previous works have used image classification problems as downstream tasks, we used in-domain pre-trained models from remote sensing images to solve semantic segmentation and object detection problems. We solved the semantic segmentation task on the DeepGlobe Land Cover Classification dataset using the DeepLabV3 model. We also used Faster-RCNN with the FPN model to solve the oil storage tanks and airplane detection. We have obtained state-of-the-art results in all dense prediction tasks using Sim-PatternNet. Although the number of training samples for Resisc45 and PatternNet are almost equal, the supervised and self-supervised pre-trained features of PatternNet have greater generalizability for solving dense prediction problems. Both datasets are multi-resolution, but assuming a uniform distribution of different spatial resolutions between pixels, the average spatial resolution of PatternNet is significantly higher than Resisc45. In addition to high-class diversity, a large sample size, and similarity with the downstream dataset, the dataset used for pre-training features from remote sensing images must also possess a high spatial resolution.